\documentclass{article}
\usepackage{spconf}
\usepackage{amsmath}
\usepackage{amssymb}
\usepackage{graphicx}
\usepackage{booktabs}
\usepackage{comment}
\usepackage{caption}
\usepackage{subcaption}
\usepackage[percent]{overpic}
\usepackage{algorithm}
\usepackage{algorithmic}

\usepackage{tikz}
\usetikzlibrary{spy}
\usepackage{pgfplots}
\pgfplotsset{compat=1.18}

\usepackage{ragged2e} 

\newcommand{\et}{\emph{et al.}}
\newcommand{\eg}{\emph{eg.,~}}
\newcommand{\ie}{\emph{ie.,~}}

\usepackage[pagebackref,breaklinks,colorlinks]{hyperref}

\title{Streaming Neural Images}

\name{Marcos V. Conde~$^{\dagger\mathsection}$ \qquad Andy Bigos~$^{\mathsection}$ \qquad Radu Timofte~$^{\dagger}$}
  
\address{
$^{\dagger}$ Computer Vision Lab, CAIDAS \& IFI, University of W\"urzburg \\
$^{\mathsection}$ Sony Interactive Entertainment}

\begin{document}
%
\maketitle
\begin{abstract}
Implicit Neural Representations (INRs) are a novel paradigm for signal representation that have attracted considerable interest for image compression. INRs offer unprecedented advantages in signal resolution and memory efficiency, enabling new possibilities for compression techniques. However, the existing limitations of INRs for image compression have not been sufficiently addressed in the literature. In this work, we explore the critical yet overlooked limiting factors of INRs, such as computational cost, unstable performance, and robustness. Through extensive experiments and empirical analysis, we provide a deeper and more nuanced understanding of implicit neural image compression methods such as Fourier Feature Networks and Siren. Our work also offers valuable insights for future research in this area.
\end{abstract}



\section{Introduction}
\label{sec:intro}

Implicit Neural Representations (INRs) allow the parameterization of signals of all kinds and have emerged as a new paradigm in the field of signal processing, particularly for image compression~\cite{tancik2020fourier, sitzmann2020implicit, dupont2021coin, strumpler2022implicit}. 
Differing from traditional discrete representations (\eg an image is a discrete grid of pixels, audio signals are discrete samples of amplitudes), INRs use a continuous function to describe the signal. Such a function maps the source domain $\mathcal{X}$ of the signal to its characteristic values $\mathcal{Y}$. For instance, it can map 2D pixel coordinates to their corresponding RGB values in the image $\mathcal{I}[x,y]$. This function $\phi$ is approximated using neural networks, thus it is continuous and differentiable. 
We can formulate this as

\begin{equation}
    \phi : \mathbb{R}^{2} \mapsto \mathbb{R}^{3} \quad \mathbf{x} \to \phi(\mathbf{x}) = \mathbf{y},
\label{eq:inr}
\end{equation}
where $\phi$ is the learned INR function, the domains $\mathcal{X} \in \mathbb{R}^{2} $ and $\mathcal{Y} \in \mathbb{R}^{3}$, the input coordinates $\mathbf{x}=(x,y)$, and the output RGB value $\mathbf{y} = [r,g,b]$. In summary, INRs are essentially simple neural networks (NN), once these networks $\phi$ (over)fit the signal, they become implicitly the signal itself. 

\begin{figure}[t]
    \centering
    \includegraphics[width=\linewidth]{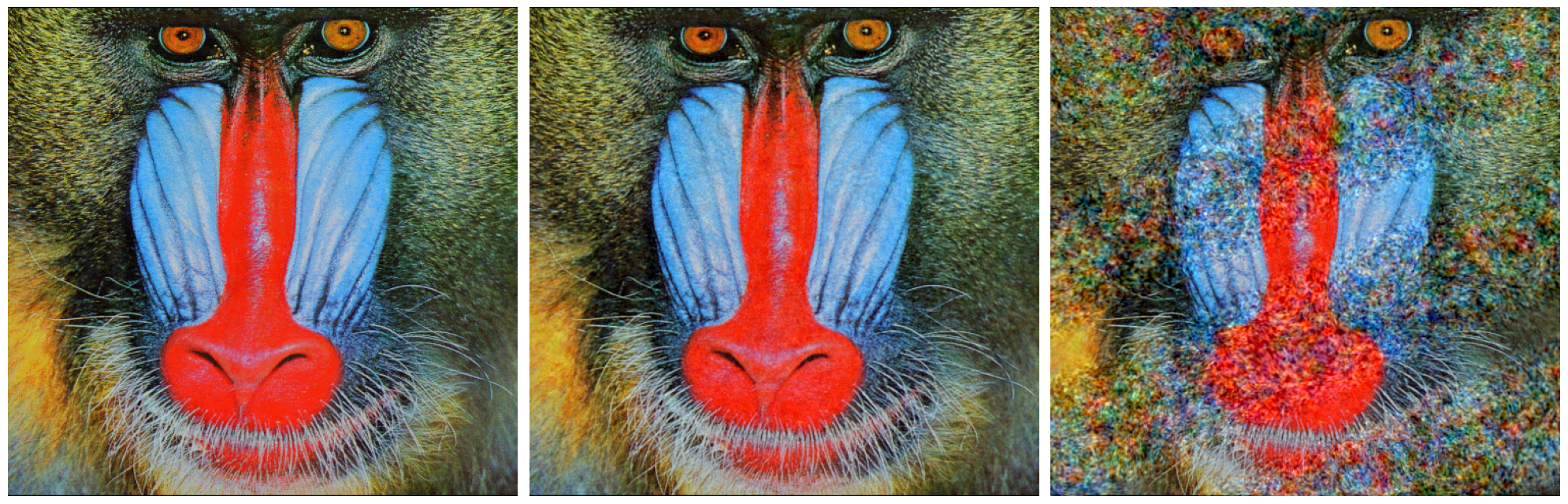}
    \vspace{-5mm}
    \caption{Exploring the behaviour of neural image representations. (Left) An image after losing one random pixel. (Mid) The corresponding implicit neural representation (INR)~\cite{sitzmann2020implicit}. (Right) The INR network after losing one random neuron.}
    \vspace{-2mm}
    \label{fig:teaser}
\end{figure}



\begin{figure*}[!ht]
     \centering
     \begin{subfigure}[b]{0.2\textwidth}
         \centering
         \includegraphics[width=\linewidth]{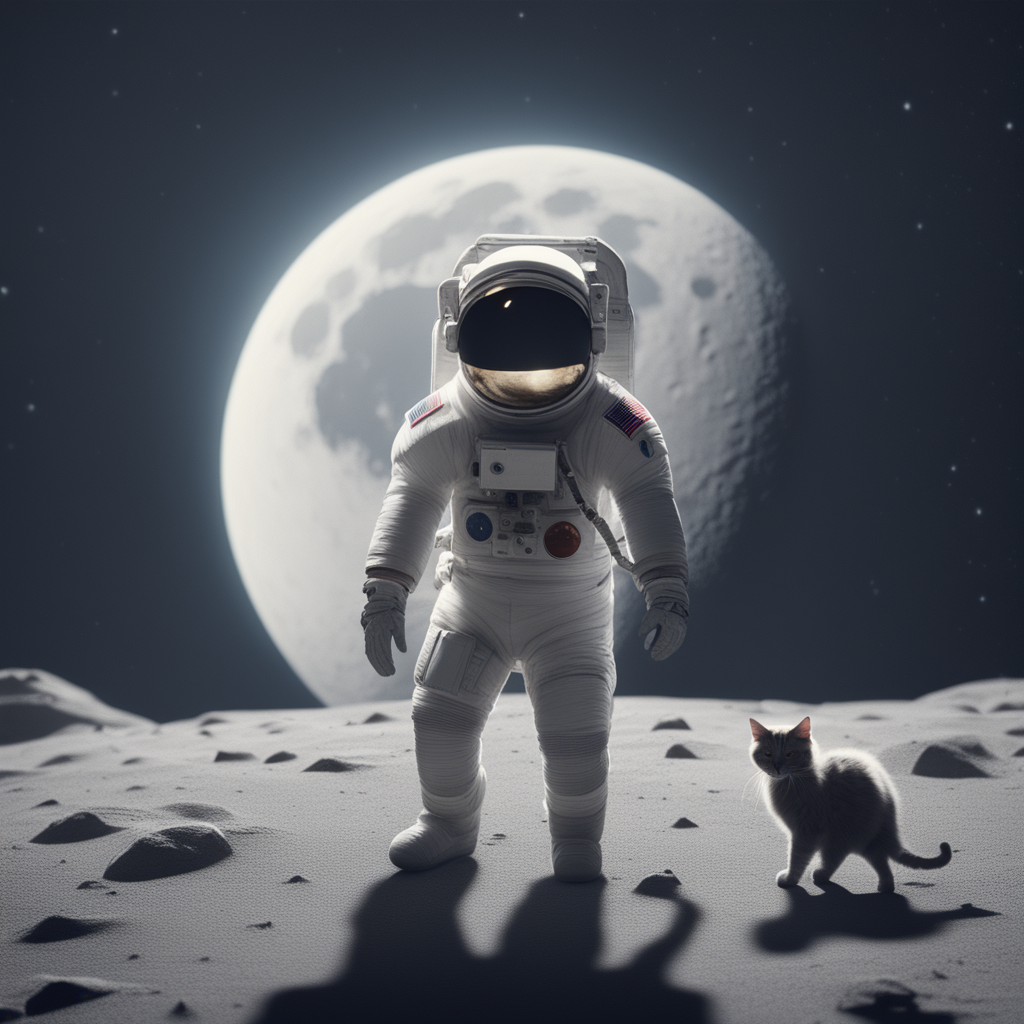}
         \put(-93,113){$\mathbf{x}= (x,y) \in [H,W]$}
         \put(-93,130){$\mathbf{x} \to (200,205,140)$}
         \put(-75,50){\Large{$\mathbf{x}$}}
         \caption{Conventional image}
         \label{fig:stream-rgb}
     \end{subfigure}
     \hfill
     \begin{subfigure}[b]{0.2\textwidth}
         \centering
         \includegraphics[width=0.85\linewidth]{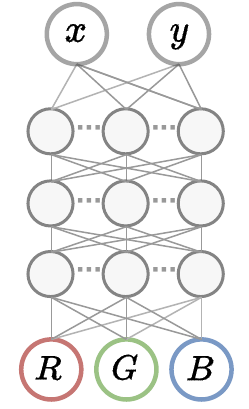}
         \caption{Coordinate-based MLP}
         \label{fig:mlp}
     \end{subfigure}
     \hfill
     \begin{subfigure}[b]{0.59\textwidth}
         \centering
         \includegraphics[width=\linewidth]{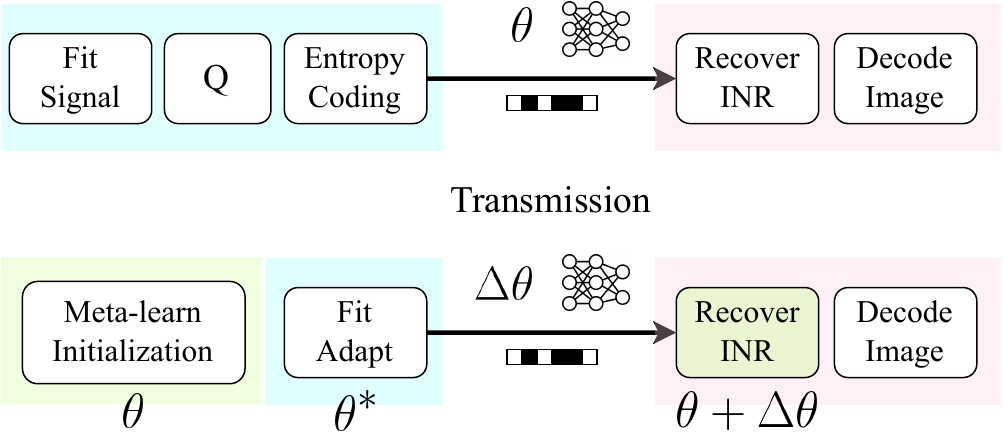}
         \vspace{0mm}
         \caption{Different variants of image streaming using INRs~\cite{strumpler2022implicit, dupont2022coin++}}
         \label{fig:stream-pipe}
     \end{subfigure}
\caption{We illustrate the general concepts around neural image representations~\cite{tancik2020fourier, sitzmann2020implicit}. We also illustrate the common frameworks for streaming images as INRs~\cite{dupont2022coin++, strumpler2022implicit}. This can be extended to other sort of signals such as audio or 3D representations.}
\label{fig:streaming}
\end{figure*}


In the context of image compression, this method offers unique adaptability thanks to its continuous and differentiable nature~\cite{dupont2021coin, strumpler2022implicit, dupont2022coin++}. 
One of the major advantages of using INRs is that they are not tied to spatial resolution. Unlike conventional methods where image size is tied to the number of pixels, the memory needed for these representations only scales with the complexity of the underlying signal~\cite{tancik2020fourier, sitzmann2020implicit}. In essence, they offer ``infinite resolution", meaning they can be sampled at any spatial resolution~\cite{sitzmann2020implicit} by upsampling the input domain $\mathcal{X}$ (\eg $[H,W]$ grid of coordinates). 

Recent works such as COIN~\cite{dupont2021coin, dupont2022coin++} and ANI~\cite{hoshikawa2024extreme} demonstrates that we can fit ``large" images (720p) using small neural networks (8k parameters) as INRs, which implies promising compression capabilities~\cite{dupont2021coin, dupont2022coin++}. These seminal works~\cite{dupont2021coin, strumpler2022implicit, hoshikawa2024extreme} show that INRs can be a better option than image codecs such as JPEG~\cite{pennebaker1992jpeg} in some scenarios (\eg at low bit-rates).

\vspace{2mm}

Considering this new paradigm, we must emphasize that an image is no longer characterized as a set of RGB pixels, but as a simple neural network (\ie an MLP). This concept poses open questions for instance, \emph{losing a pixel in an image is well-understood, but what is the equivalent in INRs? What happens if the network loses one neuron?} -- See Figure~\ref{fig:teaser}. 

\vspace{2mm}

In this work we explore in depth the limitations of INRs for image compression and streaming. We \emph{analyze major limitations} such as the volatility and stochastic nature of these neural networks, their complexity, and great sensitivity to hyper-parameters. We also introduce a \emph{novel analysis} of the robustness of these neural networks, which has important implications in the context of image transmission (Fig.~\ref{fig:streaming}). 

\begin{figure}[!ht]
    \centering
    \includegraphics[width=\linewidth]{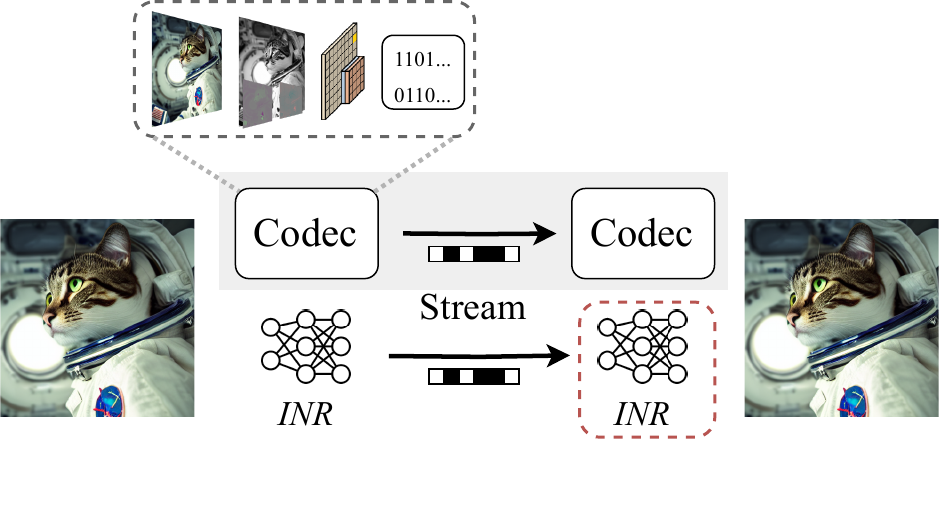}
    \put(-100,8){Do not wait!}
    \vspace{-3mm}
    \caption{Image streaming using (top) traditional image representations and codecs~\cite{pennebaker1992jpeg}, (bot.) our method, SPINR, based on implicit neural image compression~\cite{sitzmann2020implicit, dupont2021coin} allows to decode the image without having the full neural network.
    }
    \vspace{-2mm}
    \label{fig:teaser_streaming}
\end{figure}

Our approach \textbf{SPINR} (\emph{Streaming Progressive INRs}) enables to solve many of those problems, and represents a more reliable approach for implicit neural image compression and transmission. In Figure~\ref{fig:teaser_streaming}, we compare two possible solutions for efficient image transmission~\cite{rippel2017realtimecomp}.

\section{Related Work}
\label{sec:relwork}

In recent years, Implicit Neural Representations (INRs) have become increasingly popular in image processing as a new method for representing images~\cite{tancik2020fourier, sitzmann2020implicit, fathony2020multiplicative}. These neural networks, usually simple Multilayer Perceptrons (MLPs), are also known as coordinate-based networks. We denote the INRs as a function $\phi$ with parameters $\theta$, defined as: 

\begin{equation} \label{eq:mlp}
\begin{split}
\phi (\mathbf{x}) = \mathbf{W}_n ( \varsigma_{n-1} \circ \varsigma_{n-2} \circ \ldots \circ \varsigma_0 )(\mathbf{x}) + \mathbf{b}_n \\
\varsigma_i (x_i) = \alpha \left( \mathbf{W}_i \mathbf{x}_i + \mathbf{b}_i \right),
\end{split}
\end{equation}

where $\varsigma_i$ are the layers of the network (considering their corresponding weight matrix $\mathbf{W}$ and bias $\mathbf{b}$), and $\alpha$ is a nonlinear activation \eg ReLU, Tanh, Sine~\cite{sitzmann2020implicit}, complex Gabor wavelet~\cite{saragadam2023wire}. Considering this formulation, the parameters of the neural network $\theta$ is the set of weights and biases of each layer. We illustrate them in Figure~\ref{fig:mlp}.

Tancik \et~\cite{tancik2020fourier} introduced fourier features as input encodings for the network, enhancing their capability to model high-frequency functions in lower-dimensional domains. Sitzmann \et~\cite{sitzmann2020implicit} presented SIREN, a periodic activation function for neural networks, specifically designed to better model complex natural signals. Based on this work, COIN~\cite{dupont2021coin, dupont2022coin++} explored the early use of INRs for efficient image compression. 

We also find other works that tackle new activation functions such as multiplicative filter networks (MFN)~\cite{fathony2020multiplicative} and Wire~\cite{saragadam2023wire}, and novel INR representations~\cite{xie2023diner, hao2022implicit, saragadam2022miner}.

In this work we will analyze the most popular (and recent) INR models: FourierNets~\cite{tancik2020fourier} (MLP with Positional Encoding), SIREN~\cite{sitzmann2020implicit}, MFN~\cite{fathony2020multiplicative}, Wire~\cite{saragadam2023wire} and DINER~\cite{xie2023diner}.

\vspace{-2mm}

\subsection{Image Transmission}

Streaming images as INRs is a novel research problem~\cite{strumpler2022implicit, dupont2022coin++, hoshikawa2024extreme}. In this context, it is fundamental to understand that the image is no longer characterized as a set of RGB pixels, but as a set of weights and biases ($\theta$ \ie the neural network itself).

We illustrate in Figure~\ref{fig:stream-pipe} the different approaches: (top) we train the neural network $\phi$ to fit the signal, next we can apply quantization (Q) and encode the parameters $\theta$. Then we can transmit the parameters, the client can recover the network, and thus reconstruct the natural RGB image. (bot.) We use a fixed meta-INR as initialization~\cite{tancik2021metalearned, lee2021meta, chen2022metatransformers}, and adjust (fine-tune) the network to the desired image, next we transmit only a residual $\Delta\theta = \theta - \theta^*$. This allows to simplify the transmitted information and make the process more efficient~\cite{strumpler2022implicit}. Finally the client recovers the INR knowing $\Delta\theta$ and $\theta$ (meta-INR), and reconstructs the natural image.

\section{Compression Experimental Results}
\label{sec:limit}

We study the different INR methods considering the following factors related to image compression and streaming:

\begin{enumerate}
    \itemsep0em 
    \item General theoretical limitations and comparison to traditional Codecs (JPEG~\cite{pennebaker1992jpeg}, JPEG2000~\cite{skodras2001jpeg}). 
    \item Unstable training and hyper-parameters sensitivity. The performance of INR methods highly varies depending on hyper-parameters, and the target signal. 
    \item Model Complexity. The design of the neural network is paramount to ensure a positive rate-distortion tradeoff.
    \item Model Robustness. We understand noise and ``losing pixels" in classical images (Fig.~\ref{fig:stream-rgb}), however, we do not find any reference on equivalent noise and ``losing neurons" in INRs (\eg Fig.~\ref{fig:teaser}). For this reason, we introduce a novel analysis on the robustness of these neural networks --- considering also the streaming scenario.
\end{enumerate}

\noindent \textbf{Dataset} In our study we use a common dataset in image processing analysis (\eg compression, super-resolution, etc), which consists of \emph{Set5} and \emph{Set14}. The images offer wide variety of colours and high-frequencies (\eg Fig.~\ref{fig:teaser}).

\vspace{2mm}

\noindent \textbf{Implementation Details}
We implement all the methods in PyTorch, using the author's implementations when available. We train all the models using the same environment with the Adam optimizer, and we adapt the learning rate for each method's requirements. We use four NVIDIA RTX 4090Ti. We repeat every experiment 10 times with different seeds. In every experiment, each model is trained for 2000 steps (or equivalent time) using the $\mathcal{L}_2$ loss~\cite{tancik2020fourier, sitzmann2020implicit} to minimize the RGB image reconstruction error $\sum_{x,y} \| \mathcal{I}[x,y] - \phi (x,y) \|_{2}^{2} , ~~\forall (x,y) \in [H,W]$.

\subsection{General Limitations}

In the domain of image compression, INRs have shown promise but also exhibit fundamental limitations. Critically, as a lossy compression method, their ability to capture high-frequency components of images is constrained by Shannon's theory~\cite{shannon}. 
Many INR approaches, even those employing Fourier features or periodic activation functions, fall short of this, particularly for images with complex textures~\cite{tancik2020fourier, sitzmann2020implicit, fathony2020multiplicative}. 

Further, although INRs are theoretically independent of resolution, practical image discretization introduces errors that escalate with increasing resolution. Finally, most INRs approaches are signal-specific \ie the neural network fits a particular image. This might imply ``long" training on GPUs.

\subsection{Comparison with traditional Codecs}

Traditional codecs like JPEG~\cite{pennebaker1992jpeg} and JPEG2000~\cite{skodras2001jpeg}, and other neural compression techniques~\cite{balle2018variational, mentzer2020high}, have several advantages over INRs when it comes to image compression:

    \emph{Efficiency \& Speed:} Traditional codecs are optimized for low computational overhead and can rapidly compress and decompress images. INRs, on the other hand, require image-specific training, and forward passes through the neural network for image reconstruction, which can be computationally intensive. Despite meta-learning~\cite{strumpler2022implicit, tancik2021metalearned, lee2021meta, chen2022metatransformers} can help to speed up the training, it is a still a limiting factor.

    \emph{Explicit Frequency Handling:} Traditional codecs use methods like Discrete Cosine Transform (DCT) to handle frequency components explicitly. This ensures more precise control over rate-distortion ratios~\cite{shannon}. INRs \emph{learn} such high-frequencies, which is more difficult and error-prone~\cite{sitzmann2020implicit, lindell2022bacon}.

    \emph{Robustness:} Traditional codecs are generally more robust to image variations (\eg noise). This aspect was not explored in depth in the INRs literature. In Section~\ref{sec:robust} we provide a novel analysis in this direction -- also related to streaming.

\vspace{2mm}

\noindent\textbf{Summary:~} while INRs offer exciting possibilities, traditional codecs currently provide a more reliable, efficient, mature and standardized approach for image compression. 

We compare INRs with codecs in Figure~\ref{fig:jpeg}. Strumpler \et~\cite{strumpler2022implicit} also proved the limited performance of INRs in comparison to traditional codecs using other datasets.

\begin{figure}[t]
    \centering
    \input{figures/kodak-bpp}
    \caption{Comparison between INR compression~\cite{ strumpler2022implicit, dupont2021coin, dupont2022coin++} and classical JPEG compression using the Kodak dataset. Best viewed in electronic version.
    }
    \label{fig:jpeg}
\end{figure}

\begin{figure}[!ht]
    \centering
    \begin{tabular}{c}
        \includegraphics[width=0.975\linewidth]{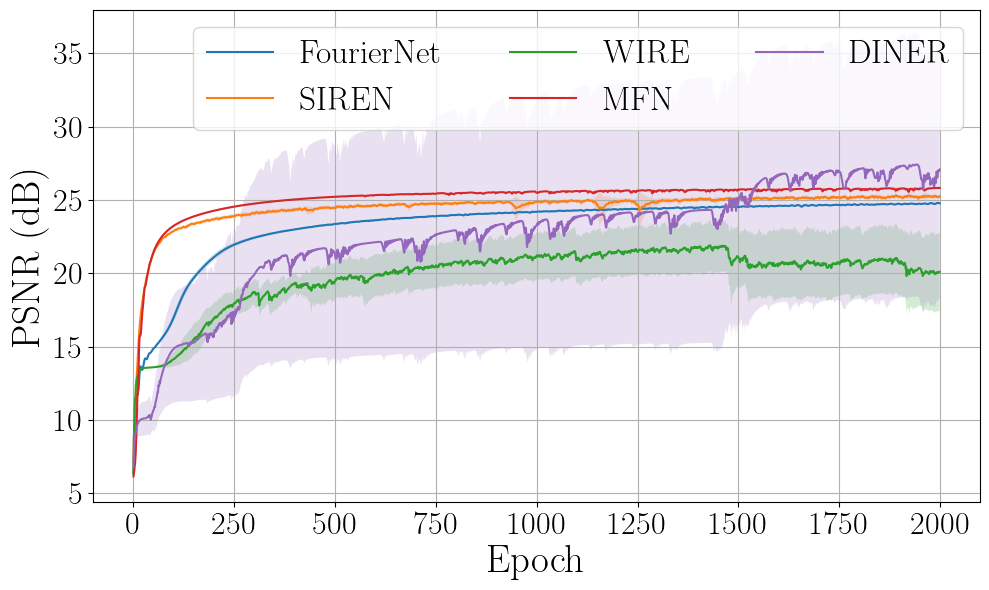}
        \put(-55,25){\includegraphics[scale=0.06]{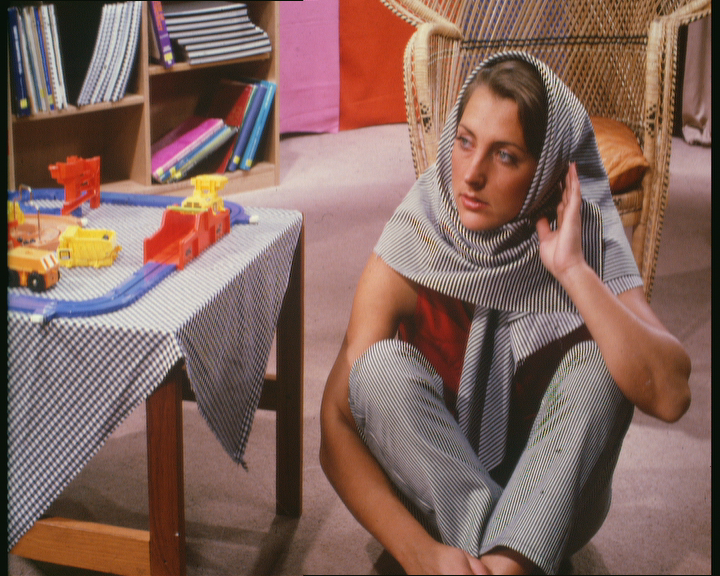}}
        \\
        \includegraphics[width=0.975\linewidth]
        {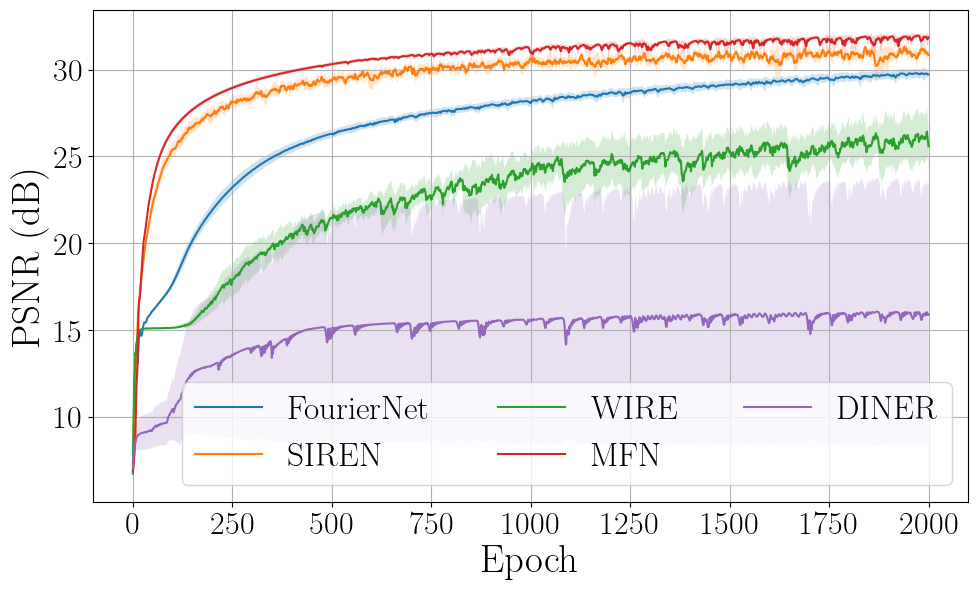}
        \put(-210,105){\includegraphics[scale=0.06]{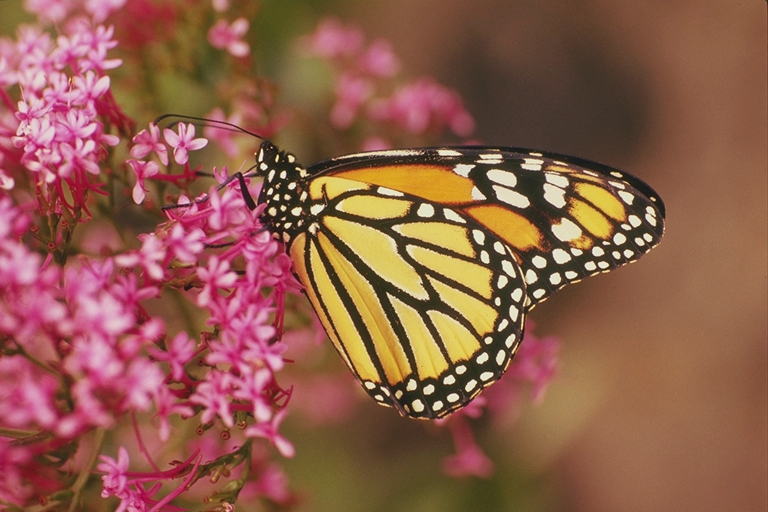}}
    \end{tabular}
    \caption{Training evolution of the different INR methods. 
    We observe high training instability for DINER~\cite{xie2023diner}. The models have $h=128, l=4$. We also show the corresponding image.
    }
    \vspace{-4mm}
    \label{fig:train}
\end{figure}

\subsection{Model Complexity}


We denote the width and depth of an INR as the number of hidden neurons ($h$) per layer, and the number of layers ($l$) respectively. The complexity of an INR (\ie number of parameters) depends on its design, $h$ and $l$. This is fundamental because large models do not offer a good compression alternative to JPEG --- as proved in~\cite{strumpler2022implicit}. For instance, we conclude from our experiments that models with $h\geq256$ and $l\geq 3$ are notably worst than JPEG and JPEG2000 in terms of rate-distortion trade-off --- we cannot even plot them in Fig.~\ref{fig:jpeg}.

Let $n$ be the number of pixels in the image. The complexity of INRs based on MLPs does not depend on the resolution of the input signal. However, in some recent methods such as DINER~\cite{xie2023diner} -based on hash mappings- the complexity does depend on the input image following an order $\mathcal{O}(n)$, where $n$ is the number of pixels in the image. This means that for an image with resolution $H, W$ the number of learnt parameters is $(H\times W \times 2) + \theta$, where $\theta$ is only the set of parameters of the MLP. Thus, these approaches offer \emph{negative} compression rate \ie the INR is bigger than the image itself.

In Table~\ref{tab:comparison} we study models with $h=256, l=2$ (highlighted with ``*") and smaller models with $h=128, l=4$ (\ie half width, double depth). We can conclude that only MLP-based approaches with ``low" complexity can be good image compressors, offering a positive compression factor (CF). For instance CF$>2$ indicates the INR is twice smaller than the natural image. See these models also in Fig.~\ref{fig:jpeg}.

Furthermore, model quantization and pruning can help to reduce the model's size while preserving high fidelity~\cite{strumpler2022implicit}.

\vspace{-2mm}

\subsection{Unstable Training and Sensitivity}
\label{sec:train}

We find that INRs are extremely sensitive to hyper-parameters, specially learning rate, $h$ and $l$. We show this volatility in Figure~\ref{fig:train} (we plot the average and confidence interval considering the 10 runs), and in Table~\ref{tab:comparison}. Recent methods, Wire~\cite{saragadam2023wire} and Diner~\cite{xie2023diner}, are specially unstable. This behaviour also varies depending on the target signal, and it is (so far) unpredictable.

\begin{table}[t]
    \centering
    \resizebox{\columnwidth}{!}{
    \begin{tabular}{l c c c c}
         \toprule
         Method & Param.~(K) & PSNR~$\uparrow$ & SSIM~$\uparrow$ & CF~$\uparrow$ \\
         \midrule
         FourierNet*~\cite{tancik2020fourier}  &  132.09 & 30.78$\pm$0.23 & 0.857 & 1.22 \\
         
         SIREN*~\cite{sitzmann2020implicit}    &  133.12 & 31.68$\pm$2.08 & 0.866 & 1.22 \\
         
         MFN*~\cite{fathony2020multiplicative} &  136.96 & 33.98$\pm$0.18 & 0.909 & 1.17 \\
         
         \midrule
         
         FourierNet~\cite{tancik2020fourier}  &  66.30 & 30.10$\pm$0.42 & 0.832 & 2.41 \\
         
         SIREN~\cite{sitzmann2020implicit}    &  66.82 & 30.95$\pm$2.31 & 0.855 & 2.40 \\
         
         MFN~\cite{fathony2020multiplicative} & 70.27  & 32.95$\pm$0.47 & 0.895 & 2.25 \\
         
         Wire~\cite{saragadam2023wire}        &  66.82 & 25.96$\pm$3.83 & 0.712 & 1.20 \\
         
         DINER~\cite{xie2023diner}            &  545.55 & 27.34$\pm$\textcolor{red}{15.5} & 0.751 & \textcolor{red}{0.32} \\
         \bottomrule
    \end{tabular}
    }
    \caption{Comparison of different INR approaches for image compression. We report the mean PSNR ($\pm$ std.) and SSIM~\cite{wang2004ssim} over 10 runs. We also show the average Compression Factor (CF$=\text{image size}  / \text{model size}$). We highlight DINER's high variability and \emph{negative} CF \ie model$>$image.
    }
    \label{tab:comparison}
\end{table}

\vspace{-2mm}

\subsection{Model Robustness}
\label{sec:robust}

\begin{figure*}[t]
    \centering
    \includegraphics[width=\linewidth]{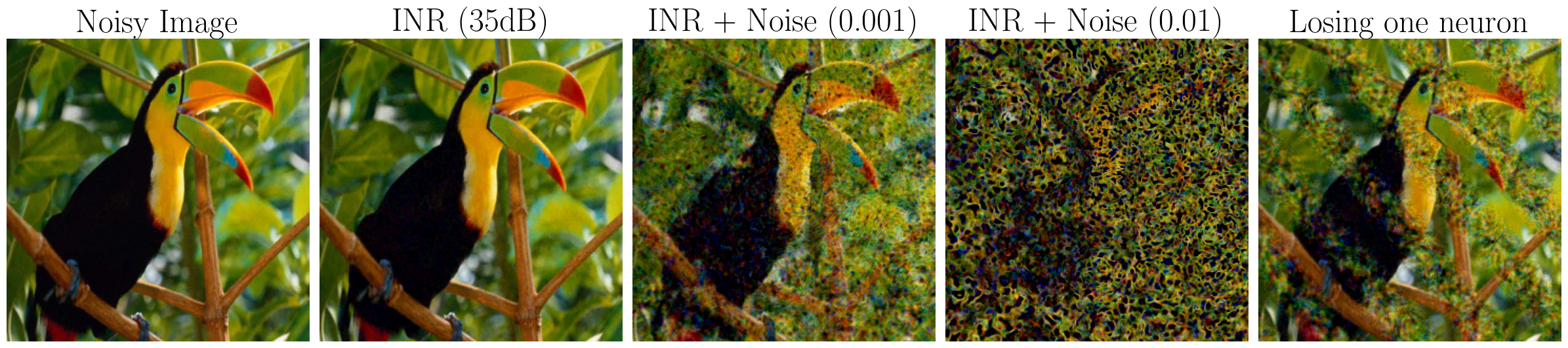}
    \caption{Illustration of robustness analysis. From left to right: noisy image ($\sigma \!=\! 0.01$), trained INR~\cite{sitzmann2020implicit}, INR with noise ($\sigma\!=\!0.001$), with noise ($\sigma\!=\!0.01$), and lost-neuron attack. In contrast to conventional images, INRs are very sensitive to noise.}
    \label{fig:attacks}
\end{figure*}

Inspired by adversarial attacks~\cite{goodfellow2014explaining}, we study how the INRs behave under information loss. For instance, noise is well-understood in natural images, however, it is unexplored in neural image representations \ie apply gaussian noise on the parameters $\theta$. We provide an example in Figure~\ref{fig:attacks}, which reveals the different behaviour of noise in both classical and implicit neural representations. These results are consistent through the whole dataset. Note that for noise levels with $\sigma\leq1e^{-4}$ the network does not suffer information loss. 

Although this is an interesting theoretical result, it is not realistic.
For this reason, we focus on the \emph{``lost-neuron"} problem \ie losing randomly some neurons of the network. Beyond theoretical interest, this is directly associated to possible packet loss during the transmission. 

We provide a study in Table~\ref{tab:loss} where we remove randomly (\ie set to 0) 1, 5, and 10 neurons from the INR, and evaluate its reconstruction performance after the corruption. We observe that losing a single neuron might imply losing most of the signal information. Noting that in the streaming use case \emph{packet loss} would imply a significantly larger number of neurons being lost. This was not considered in any previous work.
We also show in Figure~\ref{fig:attacks} the impact of losing a single neuron. We believe these results serve as the first baseline to interpretability and adversarial robustness in INRs.

\begin{table}[t]
    \centering
    \begin{tabular}{l c c c c}
        \toprule
         Method  & Base & L@1 & L@5 & L@10 \\
         \midrule
         FourierNet~\cite{tancik2020fourier}  & 30.10 & 28.50 & 24.22 & 21.32 \\
         SIREN~\cite{sitzmann2020implicit}    & 30.95 & 23.94 & 18.99 & 16.13 \\
         MFN~\cite{fathony2020multiplicative} & 32.95 & 29.94 & 25.98 & 23.74 \\
         \bottomrule
    \end{tabular}
    \caption{Robustness study to losing $k$-neurons (L@k). We report the mean PSNR (dB) over 10 trials for each setup. As we show in Fig.~\ref{fig:attacks} the performance decays notably.}
    \vspace{-2mm}
    \label{tab:loss}
\end{table}


\section{SPINR: Streamable Progressive INR}
\label{sec:ours}

Considering the previous analysis of performance and limitations of INRs, we are interested in an INR method with the following properties: (i) robustness against information loss \ie the model allows to reconstruct the RGB image even if part of the netowork $\theta$ is lost. (ii) progressive image representation \ie we can reconstruct the RGB image by stages~\cite{cho2022streamable}.

We propose SPINR, Streaming Progressive INRs, to achieve these desired properties. As a baseline, we will use SIREN~\cite{sitzmann2020implicit} following previous works~\cite{dupont2021coin, strumpler2022implicit}.

We define the neural network as a MLP with $n$ layers and $h$ neurons per layer. Strumpler~\et~\cite{strumpler2022implicit} studied the behaviour of varying the $h$ and $n$ to find a good compression rate. Following this work, we choose $n=4$ and $h=128$.

For simplification, we refer to each layer (including its weights and biases) as $L_i$. Our INR is then the application of $n=4$ hidden layers, and the input-output mappings ($L_0$ and $L_5$). We can formulate it as  $\mathbf{y} = L_5 \circ \dots \circ L_1 \circ L_0 (\mathbf{x})$.

First, the layer $L_0$ projects the 2-dimensional coordinates into the $h$-dimensional hidden space. At the end, $L_5$ maps from $h$ to the 3-dimensional RGB values. These two layers represent the smallest possible connection or network.

Our method consists on a multi-stage training to learn meaningful image representations by stages. This is similar to deep networks with stochastic depth~\cite{huang2016deep}. We describe the training process in Algorithm~\ref{algorithm}. 

\begin{algorithm}[t]
\caption{SPINR Multi-stage Training}
\begin{algorithmic}[1]

\REQUIRE Model $\phi$ with $n=4$ hidden layers ($L_0, \ldots, L_5$)

\REQUIRE Input $x$, Ground Truth $y$, Stages $S=[1,n+1]$, Number of optimization steps $N$

\REQUIRE List of forward pass layers $C=[L_0, L_5]$

\FOR{each stage $s$ in $S$}
    \STATE Initialize all model parameters as frozen
    \STATE \textbf{if}~($s=1$) Set $C$ as trainable \textbf{else} $L_s$ as trainable
    \FOR{$i=1$ to $N$}
        \STATE $\hat{y} \leftarrow$ Forward propagate $x$ through active layers $C$
        \STATE Compute and backpropagate loss between $\hat{y}$ and $y$
        \STATE Update trainable parameters $L_s$ using optimizer
    \ENDFOR
    \STATE Add $L_s$ into C (Add new learned connection)
\ENDFOR

\RETURN Trained model $\phi$

\end{algorithmic}
\label{algorithm}
\end{algorithm}

Considering $n+1$ stages, in stage 1 we train the smallest possible mapping: $\mathbf{y} = L_5 \circ L_0 (\mathbf{x})$. In the next stage, we freeze the previously trained layers (\eg [$L_0, L_5$]) and train the next connection $\mathbf{y} = L_5 \circ L_1 \circ L_0 (\mathbf{x})$, updating only $L_1$ from the loss backpropagation. We repeat this process until we train $L_n$ having as frozen layers [$L_0, ..., L_(n-1), L_5$]. This allows the model to learn the $\mathbf{x} \to \mathbf{y}$ mapping considering different connections (forward pass) within the INR.

\begin{figure*}[t]
\centering
\includegraphics[width=0.95\linewidth]{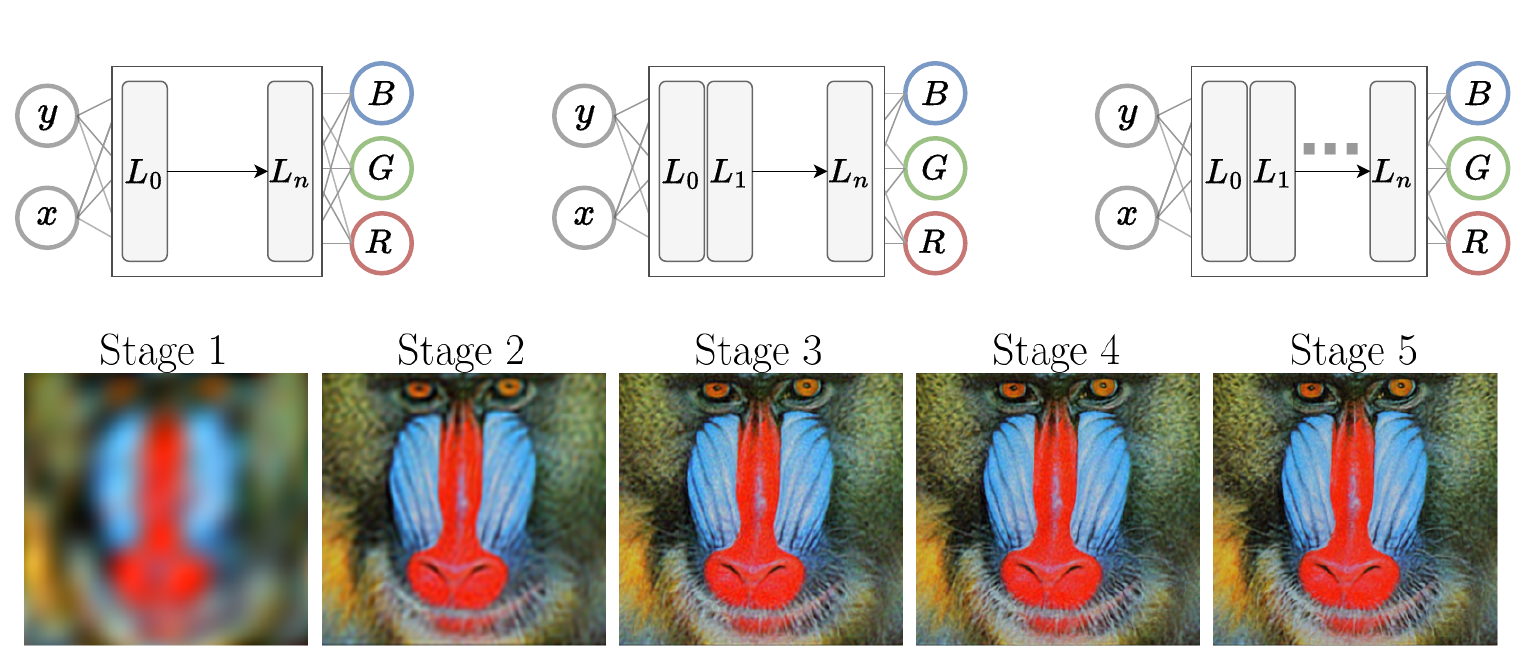}
\put(-445,195){\large $S=1 ~~(30\%)$}
\put(-270,195){\large $S=2 ~~(50\%)$}
\put(-105,195){\large $S=5 ~~(100\%)$}
\caption{We illustrate our approach SPINR (Streaming Progressive INRs). The method allows to transmit the INR neural network in stages, similar to BACON~\cite{lindell2022bacon}. This allows to reconstruct the natural image without waiting for the complete network $\theta$. Moreover, if the parameters of one layer are corrupted (\eg due to packet loss), the model can still produce an accurate image.}
\label{fig:ours}
\end{figure*}

\subsubsection*{Robustness and Progressive Decoding}

Since we learn how to map from any layer $L_i$ to the output, we do not require all the layers in the model to reconstruct the RGB image. This makes the model extremely robust to possible corruptions due to packet loss \ie losing parameters from layers. This can be seen also as layer-wise redundancy \ie if a layer is corrupted, we can still reconstruct the RGB image from the other layers. Moreover, this allows us to recover the image in a progressive manner \ie we can reconstruct it based on the received partial $\theta$, and update it when we receive additional layers. This progressive RGB reconstruction is illustrated in Figure~\ref{fig:ours}, we can appreciate how the image quality improves when we use more layer connections.

Finally, a consequence of this method is the ability to adapt the bit-rate during the transmission, for instance, we can transmit more or less information (layers) depending on the bandwidth and influence the quality of the resultant RGB.

Note that model quantization and pruning can help to improve the compression rate while preserving high fidelity~\cite{strumpler2022implicit}.

\begin{table}[t]
    \centering
    \begin{tabular}{l c c c c}
         \toprule
         Method & Params. & PSNR & L@5 & L@10 \\
         \midrule
         
         FourierNet~\cite{tancik2020fourier}  &  66.30 & 29.80 & 23.12 & 20.24 \\
         
         SIREN~\cite{sitzmann2020implicit}    &  66.82 & 28.26 & 20.00 & 15.78 \\

         \midrule

         \textbf{SPINR} (\emph{Ours}) &  66.82 & 28.32 & \textbf{24.01} & \textbf{21.67} \\

         \bottomrule
    \end{tabular}
    \caption{Comparison of different INR approaches for image compression. We report the average PSNR over 10 runs. We also show the robustness to losing $k$-neurons (L@k). Our method SPINR offers consistent performance and robustness
    }
    \vspace{-2mm}
    \label{tab:spinr}
\end{table}

\subsection{Experimental Results}
\label{sec:exp-spinir}

We use the well-known \emph{set14} dataset used in image processing analysis. For each image, we fit different INR neural networks~\cite{tancik2020fourier, sitzmann2020implicit, fathony2020multiplicative}, and we report the average reconstruction PSNR over 10 independent runs with different seeds. We report these results in Table~\ref{tab:spinr}. We can see that SPINR achieves similar performance as the other methods.

Following these experiments, and inspired by adversarial attacks~\cite{goodfellow2014explaining}, we study how the INRs behave under information loss. We focus on the \emph{``lost-neuron"} problem \ie losing randomly some neurons of the MLP, which is directly associated to possible packet loss during the transmission. 

We provide the results in Table~\ref{tab:spinr} where we remove randomly (\ie set to 0) five and ten neurons from the INR, and evaluate its reconstruction performance (also in terms of PSNR) after the corruption. We observe that losing a few neurons might imply losing most of the signal information. Noting that in real streaming scenarios, packet loss would imply a significantly larger number of parameters being lost.
Our model is more robust to losing information thanks to the proposed multi-stage training (see comparison with SIREN~\cite{sitzmann2020implicit}). Even if various layers are corrupted, the model can still leverage the information from the others.


\subsection{Discussion}
\label{sec:discussion}

Based on our experiments we can conclude that the performance of the INR methods is highly volatile, and also varies depending on the target image. However, there is no theoretical or practical way of predicting \emph{ a-priori} which method fits best the signal. This lack of predictability is significant drawback when compared to traditional encoding methods.

\textbf{When are INRs a good option?}
As we previously discussed, only low-complexity INRs offer a competitive rate-distortion trade-off in comparison to traditional codecs~\cite{skodras2001jpeg}. Tiny models (\eg $h\in\{16,32, 64, 128\}$, $l\in[1,4]$) allow to fix the bitrate (information) while optimizing for high fidelity.

Considering this, training \emph{(offline)} such models enough iterations ($\geq5000$ steps), and even repeated times, would allow to derive an INR as an optimized image-specific compressor. Nevertheless, in some cases JPEG2000~\cite{skodras2001jpeg} and advance neural compression~\cite{balle2018variational, mentzer2020high} will still be superior. Novel works such as ANI~\cite{hoshikawa2024extreme} also allow adaptive bit-rate INRs.

\vspace{-2mm}

\section{Conclusion}
\vspace{-2mm}
We provide a complete review on Implicit Neural Representations (INRs) for image compression and streaming, and we introduce a novel robustness analysis of INRs. Our method SPINR improves the robustness of the neural network to packet loss, allows progressive decoding of the compressed image, and adaptive bit-rate. Our work offers a more nuanced understanding of implicit neural image compression, providing valuable insights for future research in this field.

\noindent\textbf{Acknowledgments} This work was partially supported by the Humboldt Foundation (AvH).


\begin{figure*}[t]
    \centering
    \begin{tabular}{c}
         \includegraphics[width=\linewidth]{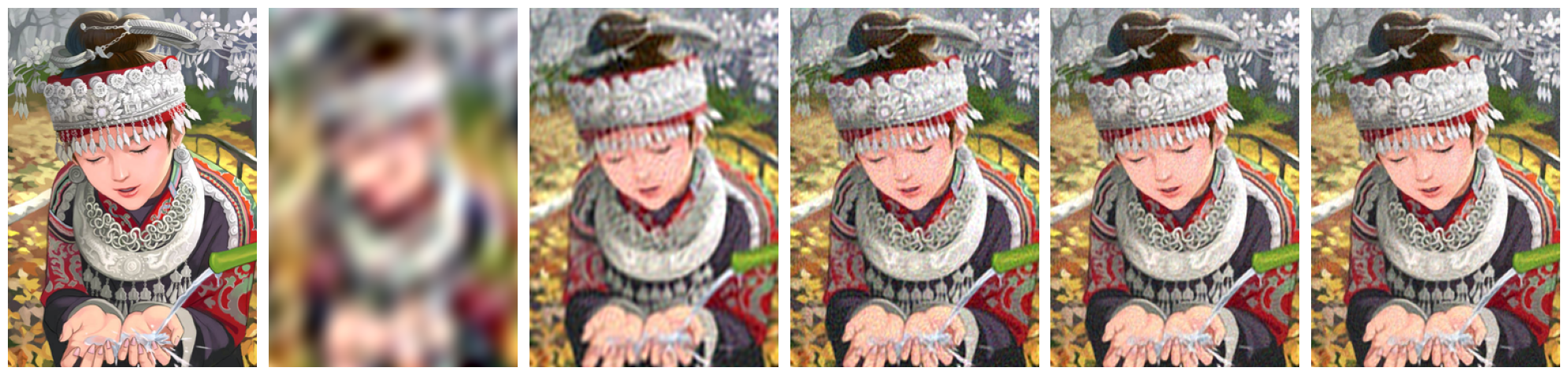} \\
         \includegraphics[width=\linewidth]{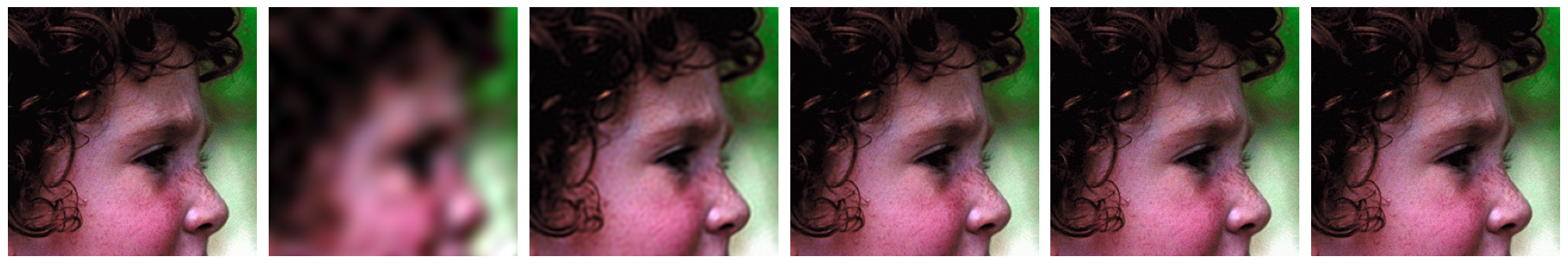} \\
         \includegraphics[width=\linewidth]{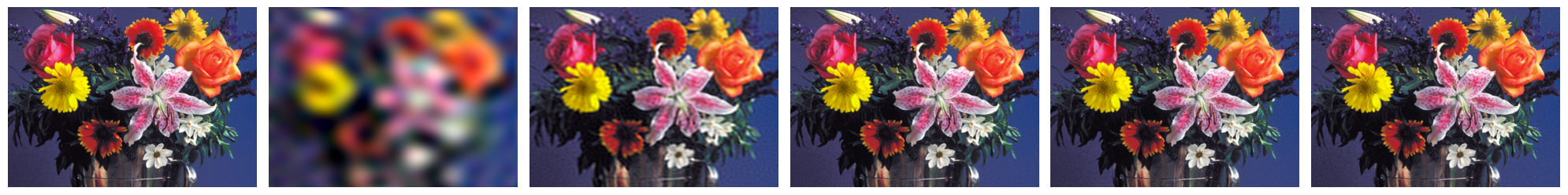} \\
         \includegraphics[width=\linewidth]{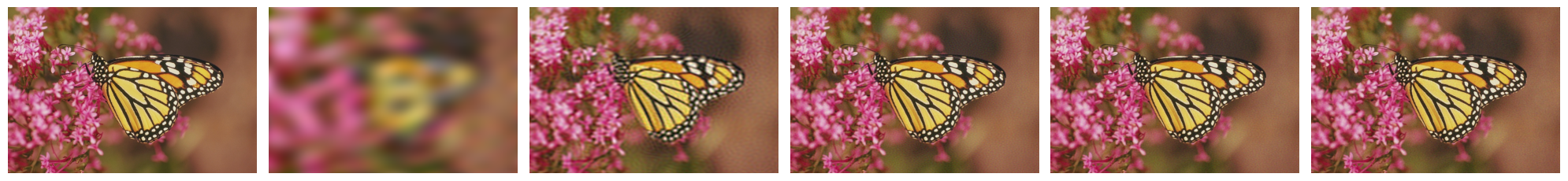} \\
    \end{tabular}
    \caption{Illustration of image transmission using the SPINR framework. From left to right: original image, stages 1 to 5. Each stage adds more details into the image by incorporating the features from an additional layer in the network. This allows to transmit the image as an INR with layer-wise redundancy for robustness, and adaptive bit-rate (\ie variable layers).}
    \label{fig:attack_samples}
\end{figure*}

{
\small
\bibliographystyle{IEEEbib}
\bibliography{refs}
}

\end{document}